\documentclass[10pt,twocolumn,letterpaper]{article}
\usepackage{cvpr}              

\usepackage{graphicx}
\usepackage{amsmath}
\usepackage{amssymb}
\usepackage{booktabs}

\usepackage[pagebackref,breaklinks,colorlinks]{hyperref}

\usepackage[capitalize]{cleveref}
\crefname{section}{Sec.}{Secs.}
\Crefname{section}{Section}{Sections}
\Crefname{table}{Table}{Tables}
\crefname{table}{Tab.}{Tabs.}

\usepackage{times}
\usepackage{graphicx}
\usepackage{newtxmath}
\usepackage{algpseudocode}
\usepackage{color}
\usepackage[accsupp]{axessibility}
\usepackage{soul}
\usepackage{multirow}
\usepackage{multicol}
\usepackage{tabularx}
\usepackage{color}
\usepackage{soul}
\usepackage{multirow}
\usepackage{tabularx}
\usepackage{float}
\usepackage{booktabs}
\usepackage{booktabs}
\usepackage[ruled]{algorithm2e}

\newcommand{\figref}[1]{Figure~\ref{#1}}
\newcommand{\equref}[1]{Equation~\ref{#1}}
\newcommand{\secref}[1]{Section~\ref{#1}}
\newcommand{\tabref}[1]{Table~\ref{#1}}

\newcommand{\subtitle}[1]{{\noindent}{\textbf{#1}}:}

\definecolor{myPurple}{rgb}{0.4, .0, .8}
\definecolor{myGreen}{rgb}{0, .8, .3}
\definecolor{myRed}{rgb}{0.8, .2, .2}
\definecolor{myOrange}{rgb}{0.8, 0.45, 0.0}
\definecolor{myBlue}{rgb}{.0, .0, 1.0}
\definecolor{myBlue2}{rgb}{.0, .0, 0.5}
\definecolor{myBlack}{rgb}{.0, .0, 0.0}

\def\cvprPaperID{4166} 
\def\confName{CVPR}
\def\confYear{2022}

\begin{document}

\title{RAGO: Recurrent Graph Optimizer For Multiple Rotation Averaging}

\author{Heng Li$^{1}$\ \ Zhaopeng Cui$^{2}$ \ \ Shuaicheng Liu$^{4}$ \ \ Ping Tan$^{1,3}$ \\
$^1$Simon Fraser University \ \ $^{2}$State Key Lab of CAD\&CG, Zhejiang University \ \ $^{3}$Alibaba XR Lab \\$^4$University of Electronic Science and Technology of China\\
{\tt\small \{lihengl,pingtan\}@sfu.ca, zhpcui@zju.edu.cn, liushuaicheng@uestc.edu.cn}
}
\maketitle

\begin{abstract}

This paper proposes a deep recurrent Rotation Averaging Graph Optimizer (RAGO) for Multiple Rotation Averaging (MRA). Conventional optimization-based methods usually fail to produce accurate results due to corrupted and noisy relative measurements. Recent learning-based approaches regard MRA as a regression problem, while these methods are sensitive to initialization due to the gauge freedom problem. To handle these problems, we propose a learnable iterative graph optimizer minimizing a gauge-invariant cost function with an edge rectification strategy to mitigate the effect of inaccurate measurements. Our graph optimizer iteratively refines the global camera rotations by minimizing each node's single rotation objective function. Besides, our approach iteratively rectifies relative rotations to make them more consistent with the current camera orientations and observed relative rotations. Furthermore, we employ a gated recurrent unit to improve the result by tracing the temporal information of the cost graph. Our framework is a real-time learning-to-optimize rotation averaging graph optimizer with a tiny size deployed for real-world applications. RAGO outperforms previous traditional and deep methods on real-world and synthetic datasets. The code is available at \href{https://github.com/sfu-gruvi-3dv/RAGO}{github.com/sfu-gruvi-3dv/RAGO}.

\end{abstract}



\section{Introduction}
\label{sec:intro}
\begin{figure}[t]
	\centering
	\scriptsize
	\includegraphics[width=0.90\linewidth]{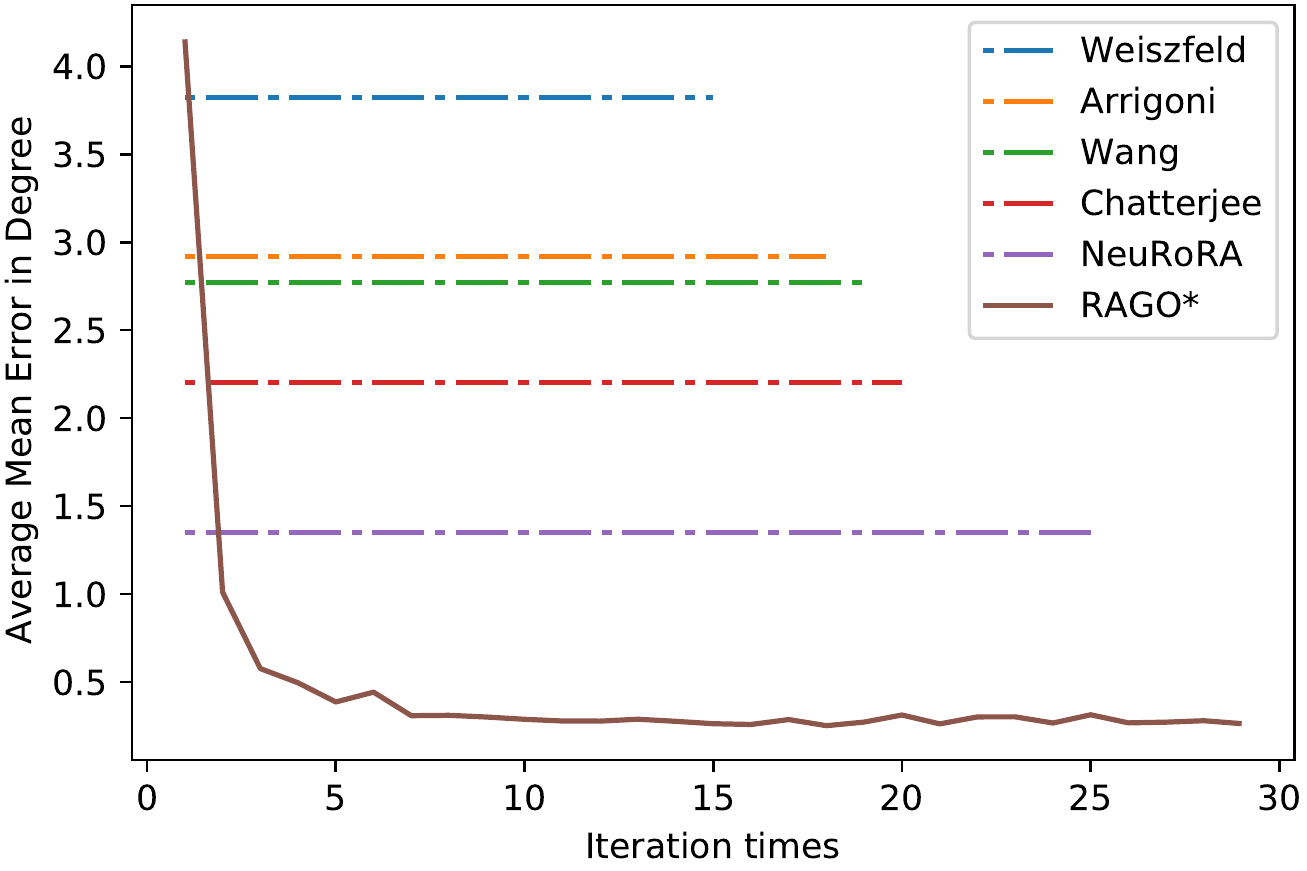}
	\caption{The average mean error of RAGO on the synthetic dataset compared with various MRA methods \cite{CVPR2011hartley,Arrigoni2018RobustSI,wang2013exact,ICCV2013,ECCV2020neurora}. The optimized results of the previous approaches are visualized as the dash lines. The vertical axis represents the average mean angular error, while the horizontal axis shows the number of iterations.}
	\label{fig:syn_mean}
    \vspace{-1em}
\end{figure}
Multiple Rotation Averaging (MRA)~\cite{hartley2013rotation,zyesil2017ASO,tron2016survey, survey2020} is a fundamental problem in 3D computer vision that aims to determine the global absolute orientations $\{\mathbf{R}_u, 1\leq u \leq N\}$ of $N$ cameras given their relative orientations $\mathbf{R}_{uv}$. It has been widely studied in 3D vision applications, \eg, global Structure from Motion (SfM)~\cite{cui2015global,clean1}, pose graph optimization in Visual Simultaneous Localization and Mapping (SLAM)~\cite{engel2017direct,mur2015orb, tang2017gslam}, and other sensor network problems~\cite{tron2009distributed,tron2008distributed}. 

MRA is usually solved by minimizing the discrepancy between the observed relative orientations $\mathbf{R}_{uv}$ and the one calculated from the estimated global orientations, i.e. $\mathbf{R}_u\mathbf{R}_v^\top$. It is a difficult classical problem with several challenges~\cite{hartley2013rotation}. Firstly, it is a highly nonlinear problem since the distance between two rotation matrices is a nonlinear function. Secondly, rotation matrices lie on the $SO(3)$ group, which requires careful parameterization and normalization during optimization.
Thirdly, there are many outliers and various noises in input relative orientations $\mathbf{R}_{uv}$, which are often computed from noisy visual correspondences across images.
These problems make the minimization objective function full of saddle points and local minimums in separate basins of attraction. Most of the time, the global optimum cannot be guaranteed. 
The MRA problem is typically solved by iterative optimization with careful initialization \cite{ICCV2013,CVPR2011hartley,IIIMA2013}. Robust cost functions with additional outlier filtering and iterative reweighting of the measurements are usually adopted, while these still do not tolerate severe corruptions and various noises.

Recently, some learning-based methods~\cite{Yang2021CVPR,ECCV2020neurora} formulate MRA as a regression problem in order to benefit from data-driven prior knowledge.
These methods rely on a good initialization, since they do not enforce any geometric constraint during inference, which may lead to an inferior result.  
Furthermore, there is a gauge freedom in MRA that prevents learning-based methods from direct end-to-end training, where $\{\mathbf{R}_{u}\}$ and $\{\mathbf{R}_{u}\mathbf{R}_0\}$ are essentially the same solution for an arbitrary rotation matrix $\mathbf{R}_0$. These learning-based methods~\cite{ECCV2020neurora,Yang2021CVPR} have to choose a root node as a reference to avoid a one-to-many mapping, which makes it hard to learn to solve the MRA problem. 

This paper presents a novel learning-based method that has the advantages of both geometrical and learning-based methods.
Specifically speaking, we decouple an MRA problem to multiple Single Rotation Averaging (SRA) problems as inspired by the traditional method in \cite{CVPR2011hartley}. An SRA problem solves the rotation matrix $\mathbf{R}_u$ from all pairwise relative rotation $\{\mathbf{R}_{uv}, v\in \mathcal{N}_u\}$, where $\mathcal{N}_u$ is the neighborhood of $u$. We construct a cost graph by computing an SRA cost function for each node independently.
We then apply a Massage Passing Neural Network (MPNN) to iteratively adjust the global camera rotations of all nodes by minimizing the SRA cost graph, resulting in an iterative optimization of the original MRA problem.
Unlike traditional methods that only can consider one-hop neighbors, MPNN has large respective fields and achieves better results. Compared to previous learning-based methods, our framework focuses on solving SRA, a much simpler and smaller problem without the gauge ambiguity, making learning easier.

In addition, in order to handle noises and outliers, we also learn to rectify the relative measurements $\mathbf{R}_{uv}$.
Our approach avoids time-consuming online refinement or edge reweighting, which makes training unstable.
In order to make it more robust and efficient, we employ a Gated Recurrent Unit (GRU) module to utilize the historical information of the previous cost graph. This module helps our optimizer to converge to a better solution. 

Experiments on real and synthetic datasets show that our method could converge to good result, even starting from random initialization, while previous methods usually required more careful initialization. As shown in \figref{fig:syn_mean}, we compare RAGO with various MRA methods on the synthetic datasets in terms of average mean angular error. RAGO outperforms these approaches after $2$ iterations.

Our contributions can be summarized as follows:
\begin{itemize}
    \vspace{-0.1cm}
    \item We present a novel end-to-end learning-to-optimize recurrent graph neural network for MRA. 
    \vspace{-0.1cm}
    \item We decouple an MRA problem to multiple SRA problems, leading to better results and learning without gauge ambiguity.
    \vspace{-0.1cm}
    \item We propose to rectify the relative orientations during optimization to handle outliers and noises.
    \vspace{-0.1cm}
	\item Our method outperforms state-of-the-art methods on multiple real and synthetic datasets.
\end{itemize}

\begin{figure*}[ht]
	\centering
	\scriptsize
	\includegraphics[width=\textwidth]{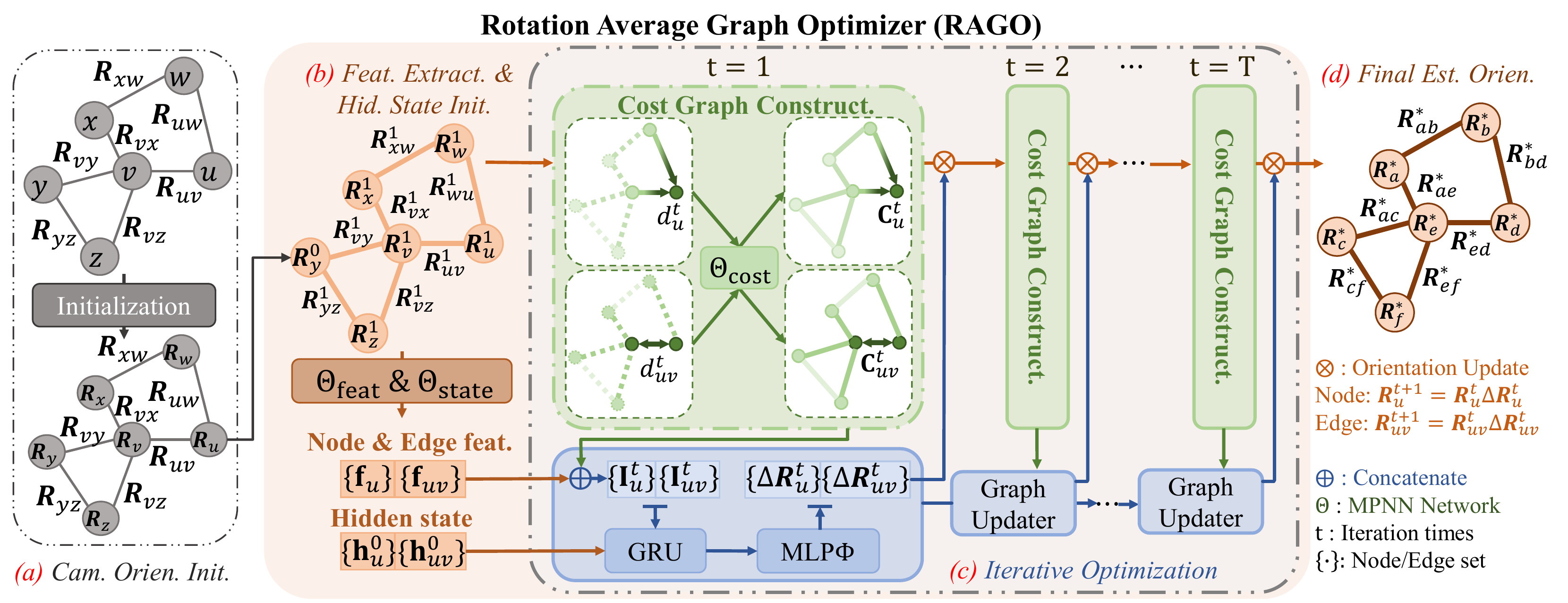}
\caption{The pipeline of our Rotation Averaging Graph Optimizer (RAGO). (a) We assign a random rotation to each camera. In (b), Two Message Passing Neural Network, $\Theta_{\text{feat}}$ and $\Theta_{\text{state}}$, extract the feature and initialize the hidden state. (c) During iterative optimization, we first build a cost graph based on the SRA objective function. Then, we employ a GRU to update the camera orientations to minimize the cost graph. (d) The camera orientations eventually converge to an optimized solution. Please refer to the main text for more details.}
	\label{fig:pipeline}
    \vspace{-1em}
\end{figure*}
\section{Related Work}
\label{sec:relatedwork}

\subtitle{Conventional MRA}
Govindu first introduced MRA with his linear motion model~\cite{govindu2001} and lie-group-based averaging~\cite{govindu2004}. More recent iterative optimization-based approaches~\cite{ICCV2013,CVPR2011hartley,IIIMA2013,Shi2020MessagePL,IIIMA2013,DISCO} introduce robust optimization strategies to reduce the influence of outliers. 
The vast majority of these algorithms were iterative and aimed to optimize a robust cost function. 
Hartley \etal~\cite{CVPR2011hartley} optimized each camera's absolute orientation using the median orientation calculated from its neighbors in each iteration using the Weiszfeld averaging algorithm.
Chatterjee \etal~\cite{ICCV2013} fine-tuned the initialization provided by a spanning tree using iterative reweighted least-squares (IRLS) minimization with an L1 loss function.
Fredriksson and Olsson \cite{duality} turn the original problem into a dual problem utilizing Lagrangian Duality and then solve it using SDP to arrive at an optimized solution. 
Numerous approaches~\cite{Arrigoni2018RobustSI,dellaert2020shonan,parra2021rotation,moreira2021rotation} are based on this pipeline for improved performance, as this approach benefits in achieving the global minimum.
Numerous approaches~\cite{Arrigoni2018RobustSI,dellaert2020shonan} are based on this pipeline for improved performance, as this approach benefits in achieving the global minimum.
Dellaert \etal~\cite{dellaert2020shonan} solves the MRA locally on $SO(3)$ and then increases the manifold dimension to start the optimization again.
Moreira \etal~\cite{moreira2021rotation} present a primal-dual method to solve MRA, inspired by in optimization algorithms with orthogonality constraints.
These approaches primarily aim to decrease the complexity of non-convex optimization. Dealing with outliers remains an open issue, as they either assume no noise or assume a specific kind of noise model.

\subtitle{Learning-based MRA} Recently, several neural network based methods\cite{ECCV2020neurora,Yang2021CVPR,huang2019learning,gojcic2020learning,yew2021learning} have been proposed. NeuRoRA \cite{ECCV2020neurora} employs a two-stage neural network architecture based on MPNN \cite{MPNN}. The first network filters outliers and rectifies relative orientations to improve the SPT-based initialization. The second stage fine-tunes the camera's orientation for a better result. MSP \cite{Yang2021CVPR}, based on NeuRoRA \cite{ECCV2020neurora}, takes appearance information as input and introduces a differentiable SPT method to achieve a robust initialization result. The initialization is further improved through non-learnable iterative edge reweighting. 
However, these approaches have to choose a node as root to enforce a unique solution, making their results sensitive to initialization.
In contrast, we regard the MRA as an optimization problem and iteratively update the variables with a message passing neural network combined with gated recurrent units to exploit temporal information.

\subtitle{Learning to Optimize}
Numerous recent publications attempt to combine the strength of neural networks with classic optimization-based methods. There are primarily two dominant directions in optimization learning. The first one \cite{amos2017optnet,agrawal2019differentiable,tang2018ba,clark2018learning} substitutes a neural network for the non-differentiable component of a traditional optimizer during end-to-end training. Other approaches \cite{qi2017pointnet,clark2018ls,chen2017learning,adler2017solving} use machine learning to update optimization variables based on the input data directly. However, all approaches need explicit formulation of the solver and are restricted to problems with easily defined objective functions. Additionally, the approaches \cite{clark2018ls,tang2018ba} must evaluate the gradient of the objective functions, which is complicated with many issues, particularly optimization on a manifold. 
Unlike these previous works, our method decouples MRA to multiple SRA problems, which is easier to learn, avoiding gradient computations.

\section{Deep Graph Optimizer for MRA}
Consider a view-graph $\mathcal{G}=\{ \mathcal{V}, \mathcal{E}\}$ where each vertex $v\in\mathcal{V}$ corresponds to an unknown absolute camera orientation and each edge $e\in \mathcal{E}$ is an observed relative orientation. The MRA problem aims to estimate a set of camera orientations $\{\mathbf{R}_u^{*}\} =\{ \mathbf{R}^*_1,\dots, \mathbf{R}^*_N\}$ that minimizes the discrepancy between the estimated and observed relative orientations, which can be formulated as:
\begin{equation}
 \{\mathbf{R}_{u}^{*}\} = \mathop{\arg\min}_{\{\mathbf{R}_u\}}\sum_{(u,v) \in \mathcal{E}} \rho ( d(\mathbf{R}_{uv}, \mathbf{R}_u\mathbf{R}_v^{\top}) ),
  \label{eq:mra_objective}
\end{equation}
where $\{\mathbf{R}^{*}_u\}$ is the set of optimized global camera orientations, $\rho(\cdot)$ is a robust cost function and $d(\cdot, \cdot)$ is the distance between two rotation matrices.
\label{sec:method}

In Single Rotation Averaging (SRA) \cite{hartley2013rotation}, a single rotation is averaged over several observed relative rotations, which can be formulated as:
\begin{equation}
    \mathbf{R}_{u}^{*} = \arg\min_{\mathbf{R}_{u}} \sum_{v\in\mathcal{N}_{u}}\rho(d(\mathbf{R}_u, \mathbf{R}_{uv}\mathbf{R}_{v}) ),
    \label{eq:sra_on_mra}
\end{equation}
where $\mathcal{N}_{u}$ is the set of neighboring nodes of $u$.
The MRA problem can be solved by iteratively solving multiple SRA problems to adjust each camera's rotation based on the orientations of its direct neighbors \cite{CVPR2011hartley}.
The overall cost decreases at each step of this procedure and therefore converges to a local minimum.

\subsection{Overview}
\label{sec:method_overview}

The overall pipeline of our framework is depicted in the \figref{fig:pipeline}. For camera orientation initialization in \figref{fig:pipeline} (a), we assign a random rotation to each camera, because our method is designed to work with random initialization.
We can also use more sophisticated initialization\cite{Yang2021CVPR, ECCV2020neurora} to enhance the robustness of our method further.
In particular, we choose CleanNet-SPT \cite{ECCV2020neurora} initialization for all real-world datasets.

In \figref{fig:pipeline} (b), two neural networks based on Message Passing Neural Network (MPNN)~\cite{MPNN}, $\Theta_{\text{feat}}$ and $\Theta_{\text{state}}$, extract local features $\{\mathbf{f}_u, \mathbf{f}_{uv}\}$ and initialize the hidden state $\{\mathbf{h}_u^0, \mathbf{h}_{uv}^0\}$ of each node and edge in the view-graph, which is introduced in \secref{sec:feat_ext} and \secref{sec:state_init}.

\figref{fig:pipeline} (c) is the iterative optimization explained in \secref{sec:iter_opt}, where we compute an SRA cost $\{d_u^t\}$ at each node and $\{d_{uv}^t\}$ at each edge from the current result at the $t$-th iteration. Then, an MPNN $\Theta_{\text{cost}}$ extracts the cost feature $\{\mathbf{C}_u^t, \mathbf{C}_{uv}^t\}$ from the cost graph $\{d_u^t, d_{uv}^t\}$. The graph updater receives the cost features $\{\mathbf{C}_u^{t}, \mathbf{C}_{uv}^t\}$, the graph features $\{\mathbf{f}_u, \mathbf{f}_{uv}\}$, and the previous hidden states $\{\mathbf{h}_u^{t-1}, \mathbf{h}_{uv}^{t-1}\}$ to generate an incremental update $\{\Delta \mathbf{R}^t\}$ to minimize the cost graph. 

After several iterations, the global camera orientation converges to an optimized solution as \figref{fig:pipeline} (d). The Algorithm~\ref{code:RAGO} demonstrate the pipeline of RAGO without alternative optimization mentioned in \secref{fig:alternate_iter}.



\subsection{Rotation Averaging Graph Optimizer(RAGO)}
\subsubsection{Graph Feature Extraction}
\label{sec:feat_ext}

We use a Message Passing Neural Network~\cite{MPNN} (MPNN) $\Theta_{\text{feat}}$ with one Edge Convolution layer as the backbone to extract the node feature $\{\mathbf{f}_{u}\}$ and edge feature $\{\mathbf{f}_{uv}\}$ of the input view-graph. We replace camera orientations on nodes with zero vectors and only extract features from observed relative orientations on edges. 
Consider an edge $e_{uv}$ with a feature $\mathbf{f}_{uv}$ connecting nodes $u$ and $v$, where the node feature is denoted by $\mathbf{f}_u$ and $\mathbf{f}_v$.
At each Edge Convolution layer, the node and edge features are updated by aggregating their neighbors' information and then passed to the next layer.
To update the edge feature, Edge Convolution concatenates the node feature and edge feature as $[\mathbf{f}_{u},\mathbf{f}_{v}, \mathbf{f}_{uv}]$. Then the concatenated feature is passed through a $3$-layer Multi-layered Linear Perception (MLP) $\Phi_{\text{edge}}$ to generate the updated edge feature $\mathbf{f}_{uv}'$.  
A node MLP $\Phi_{\text{node}}$ then updates the node feature $\mathbf{f}'_{u}$ by aggregating the adjacent updated edge features.
The structure inside an Edge Convolution is as follows:
\begin{equation}
\begin{split}
     \mathbf{f}^{'}_{uv} & = \Phi_{\text{edge}}([\mathbf{f}_{u},\mathbf{f}_{v},\mathbf{f}_{uv}]), \\
     \mathbf{f}^{'}_{u}  & = \Phi_{\text{node}}(\text{mean}(\{\mathbf{f}^{'}_{uv}, v\in \mathcal{N}_{u}\})),
\end{split}
\label{eq:edge_conv_equ}
\end{equation}
where $\mathcal{N}_{u}$ is the neighbor node set of the node $u$.
Finally, we apply a $3$-layer MLP to nodes and edges from the final Edge Convolution layer to get a feature of specified dimension on both nodes as $\{\mathbf{f}_{u}\}$ and edges as $\{\mathbf{f}_{uv}\}$. Please refer to the supplementary material for more details.

\subsubsection{Hidden state Initialization}
\label{sec:state_init}
The Gated Recurrent Unit (GRU) module in our graph updater introduced in \secref{sec:graph_updater} requires a hidden state for each node and edge in the view-graph to utilize temporal information during iteration. We generate the initial hidden states $\{\mathbf{h}_u^{0}\}$ on nodes and $\{\mathbf{h}_{uv}^{0}\}$ on edges by passing the view-graph to another MPNN $\Theta_{\text{state}}$, which has the same structure and input as $\Theta_{\text{feat}}$. Finally, The hidden states $\{\mathbf{h}_u^0, \mathbf{h}_{uv}^0\}$ is mapped to $(-1,1)$ by the tanh function. Similar to the graph feature extraction, the orientations on nodes are replaced with zeros vectors.

\subsubsection{Iterative optimization}
\label{sec:iter_opt}

\subtitle{SRA Cost Graph Construction}
\label{sec:define_cost}
At the heart of our proposed framework is the construction of the cost graph. Here we define a cost on each node and each edge.

Conventional optimization-based methods for MRA usually use \equref{eq:mra_objective} as the objective function. 
However, it is difficult to enforce the minimization of this objective function in learning-based methods. Compared with solving MRA directly, SRA is a simpler problem and easier to learn for a neural network. Thus, we decouple the MRA problem into multiple SRA problems, and compute an SRA cost for each node as,
\begin{equation}
    d_{u}^{t} = \frac{1}{|\mathcal{N}_u|}\sum_{v\in \mathcal{N}_u}||\mathbf{R}_{u}^{t} - \mathbf{R}_{uv}\mathbf{R}_{v}^{t}||_1,
    \label{eq:cost_on_node}
\end{equation}
where $||.||_1$ is the L1 norm, $\mathcal{N}_{u}$ is the neighbor node set of node $u$.

Outlier rejection during optimization is non-trivial too. 
Previously, many methods reweight the edge during optimization. We find that reweighting the edge makes training unstable. In contrast,
we introduce a relaxing parameter $\mathbf{R}^{t}_{uv}$ on each edge to mitigate the influence of outliers and noisy orientations during optimization.
In particular, for each input relative orientation $\mathbf{R}_{uv}$, we compute the estimated relative orientation from global camera orientations as $\mathbf{R}_{u}^{t}\mathbf{R}_{v}^{t\top}$. We then estimate a rectified relative orientation $\mathbf{R}_{uv}^{t}$ that is close to both $\mathbf{R}_{uv}$ and $\mathbf{R}_{u}^t\mathbf{R}_{v}^{t\top}$. If a measurement $\mathbf{R}_{uv}$ is an outlier, the rectified rotation $\mathbf{R}_{uv}^{t}$ will be far from the input rotation $\mathbf{R}_{uv}$.
The SRA cost function on the edge is then defined as:
\begin{equation}
    d_{uv}^{t} = ||\mathbf{R}_{uv}^{t} - \mathbf{R}_{u}^{t}\mathbf{R}_{v}^{t\top}||_1 + ||\mathbf{R}_{uv}^{t} -\mathbf{R}_{uv}||_1.
    \label{eq:cost_on_edge}
\end{equation}
Finally, an MPNN $\Theta_{\text{cost}}$ with three Edge Convolution layers extracts cost features $\{\mathbf{C}^t\}$ from the cost graph $\{d^t\}$. The cost feature at a node or edge has information on all of its $3$-order neighbors, because it is updated three times by the Edge Convolutions, which leads to better convergence in our iterative optimization.
Notice that the choice of distance $d(\cdot, \cdot)$ and robust functions $\rho(\cdot)$ is trivial in RAGO since the cost eventually maps to a feature space. 

\subtitle{Recurrent Graph Updater}
\label{sec:graph_updater}
Our graph updater includes two GRUs~\cite{cho2014learning} to update the global camera orientations on the nodes and the rectified relative orientations on edges, respectively. The GRUs can efficiently utilize the information in the previous iteration steps for better optimization.
We concatenate the current cost feature on the node (edge) $\mathbf{C}^{t}$, current orientations $\mathbf{R}^{t}$, and the graph local feature $\mathbf{f}$ to create an input $\mathbf{I}^{t}=[\mathbf{C}^{t},\mathbf{R}^{t},\mathbf{f}]$ for each iteration.
GRUs receive previous hidden states $\{\mathbf{h}^{t-1}\}$ and the current input $\{\mathbf{I}^{t}\}$, then outputs the current hidden states $\{\mathbf{h}^{t}\}$. Then, the incremental update rotation $\Delta\mathbf{R}^{t}$ is predicted from the hidden state $\{\mathbf{h}^t\}$ by an MLP:
\begin{equation}
    \begin{aligned}
    \mathbf{h}^{t} &= \text{GRU}(\mathbf{h}^{t-1}, \mathbf{I}^{t}), \\
    \Delta\mathbf{R}^{t} &= \Phi(\mathbf{h}^{t}).
    \end{aligned}
    \label{eq:updater}
\end{equation}
The estimated and rectified orientations on nodes and edges are then updated as:
\begin{equation}
\mathbf{R}^{t+1}_{u} = \mathbf{R}^{t}_{u}\Delta\mathbf{R}^{t}_{u},\; \; \; \mathbf{R}^{t+1}_{uv} = \mathbf{R}^{t}_{uv}\Delta\mathbf{R}^{t}_{uv}.
\end{equation}

With this rotation averaging graph optimizer, starting from an initial guess, the orientations on the view-graph are refined by the optimization iterations and eventually converge to optimized camera orientations and relative orientations, $\mathbf{R}^*_{u} \leftarrow \mathbf{R}^{t}_{u},\; \; \; \mathbf{R}^*_{uv} \leftarrow \mathbf{R}^{t}_{uv}$.

\begin{figure}[t]
	\centering
	\scriptsize
	\includegraphics[width=\linewidth]{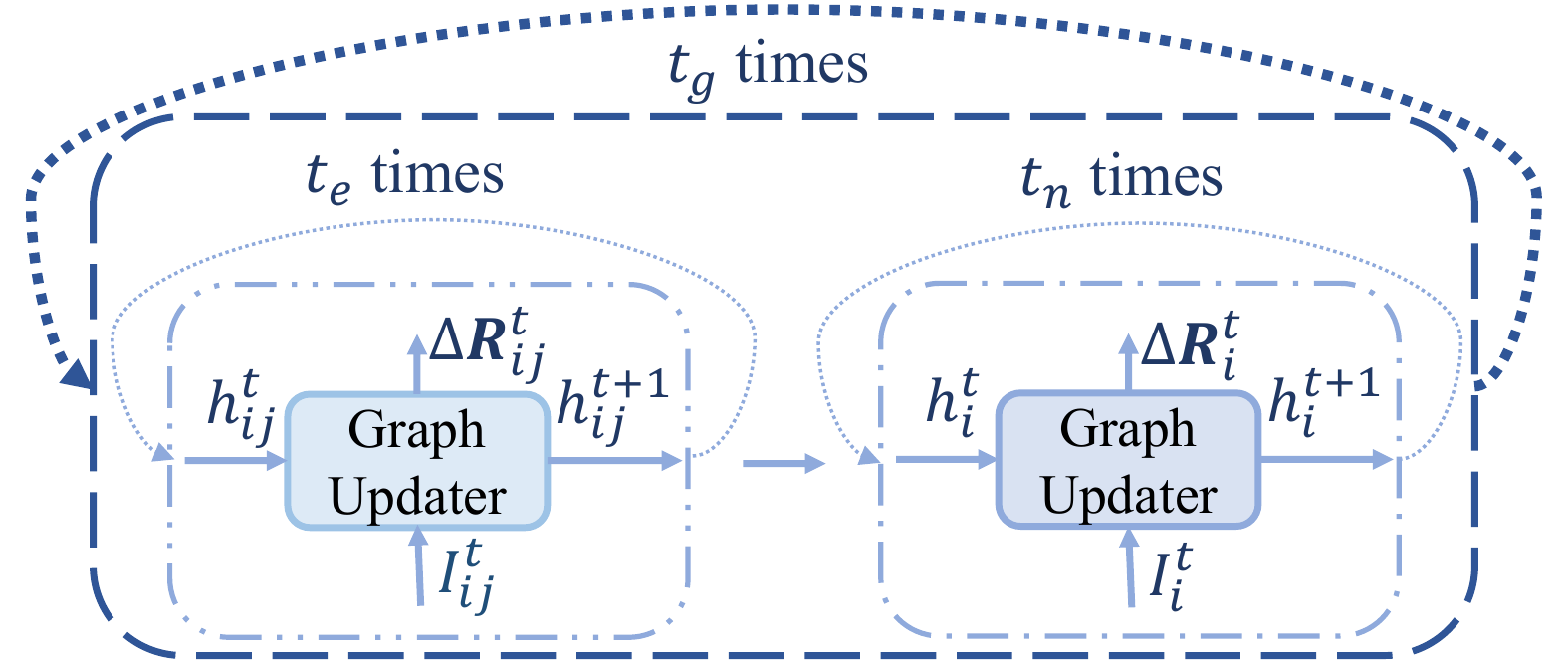}
	\caption{Alternative Optimization. RAGO first refines the relative orientations for $t_e$ times, and then optimizes the estimated camera orientations for $t_n$ times. The orientations on edges and nodes will iterative $t_e\times t_g$ and $t_n\times t_g$ times in total, respectively.}
	\label{fig:alternate_iter}
    \vspace{-1em}
\end{figure}
\subsubsection{Alternative Optimization}

Although we can optimize the variables on nodes and edges simultaneously, we find that the rectified relative rotation converges faster than the estimated camera orientation. Thus, as shown in the \figref{fig:alternate_iter}, we iteratively optimize edge and node in turns. For all of our experiments, we fix the number of iterations for graph optimization $t_g$, edge optimization $t_e$, and node optimization $t_n$ during training.

\subsection{Training Loss}
\label{training}



\label{sec:training_loss}
We train our graph optimizer in a supervised manner with ground-truth camera orientations. 
Different from the previous learning-based methods \cite{ECCV2020neurora,Yang2021CVPR}, we do not define a loss on the absolute camera rotation to enforce a unique result. We only use ground-truth relative orientations to supervise our graph optimizer:

\begin{equation}
\begin{aligned}
    \mathit{L}_{\text{opt}} = &\frac{1}{|\mathcal{E}|}\sum_{i=1}^{T_n}\sum_{(u,v)\in \mathcal{E}} \gamma^{T_n-i} ||\mathbf{R}^{i}_{u}\mathbf{R}^{i\top}_{v} - \bar{\mathbf{R}}_{uv}||_{1} +\\
    &\frac{1}{|\mathcal{E}|}\sum_{i=1}^{T_e}\sum_{(u,v)\in \mathcal{E}} \gamma^{T_e-i} ||\mathbf{R}^{i}_{uv} - \bar{\mathbf{R}}_{uv}||_{1},\\
\end{aligned}
\label{eq:loss_refine}
\end{equation}
where $\bar{\mathbf{R}}_{uv}$ is the ground truth relative orientation, $\gamma$ is a discounting factor and $T_n$ and $T_e$ are the total number of optimization iterations for nodes and edges, where $T_e=t_g\times t_e$, and $T_n=t_g\times t_n$. 
    \begin{algorithm}
    \footnotesize
        \SetAlgoLined
        \SetKwInOut{Input}{Input}\SetKwInOut{Output}{Output}
        \Input{$\mathcal{G}=\{\mathcal{V}, \mathcal{E}\}, \{\mathbf{R}_{uv}\}|(u,v)\in \mathcal{E}$}
        \Output{Global absolute rotations $\mathbf{R}_u$ for $u \in \mathcal{V}$}
        \BlankLine
        \emph{Initialization:}\
        
        \nl $R^1_{u} \gets \text{rand}(SO(3)), \forall u\in \mathcal{V}$ \Comment{Set initial to a random rotation}
        
        \nl $R^1_{uv} \gets \mathbf{R}_{\mathbf{1}}, \forall uv\in \mathcal{E}$ \Comment{Set initial to identity matrix}
        
        \nl $\{\mathbf{f}\} \gets \Theta_{\text{feat}}(\mathbf{R}_{uv})$ \Comment{Extract feature from relative rotations}
        
        \nl $\{\mathbf{h}^{0}\} \gets \Theta_{\text{state}}(\mathbf{R}_{uv})$ \Comment{Initialize hidden state for GRU}
        
        
        
        \BlankLine
        \emph{Iterative Optimization:}\
        
        \nl $t \gets 1$ 
                                                    
        \While{$(t \leq T)  $}{

                   \BlankLine
        \emph{Building Cost Graph:}\
           
           \nl $\{\mathbf{d}^{t}_{u}\} \gets \text{Equation~\ref{eq:cost_on_node}}$ \Comment{Computing SRA cost}
           
           \nl $\{\mathbf{d}^{t}_{uv}\} \gets \text{Equation~\ref{eq:cost_on_edge}}$ \Comment{Computing edge cost}


          \nl $\{\mathbf{C}^{t}\} \gets \Theta_{\text{cost}}(\{\mathbf{d}^{t}\})$ \Comment{Extracting cost feature}
           
                   \BlankLine
        \emph{Updating Rotation:} \

            \nl $\mathbf{I}^{t} = [\mathbf{C}^{t}, \mathbf{R}^{t},\mathbf{f}]$ \Comment{Creating input feature}

            
            \nl $\{\Delta\mathbf{R}^{t}\},\{\mathbf{h}^{t}\} \gets \text{Equation~\ref{eq:updater}}$ \Comment{Computing updating rotation}

           \nl $\mathbf{R}_{u}^{t+1} = \mathbf{R}^{t}_{u}\Delta\mathbf{R}^{t}_{u}$ \Comment{Updating camera rotation}
           
           \nl $\mathbf{R}_{uv}^{t+1} = \mathbf{R}^{t}_{uv}\Delta\mathbf{R}^{t}_{uv}$\Comment{Updating rectification matrices}

           \nl $t \gets t + 1 $ 
 
           }
        \caption{RAGO Inference}
        \label{code:RAGO}
        
    \end{algorithm}

\section{Experiments}
\label{sec:experiments}

\subtitle{Synthetic dataset} We evaluate on the public synthetic dataset~\cite{ECCV2020neurora}. This dataset is generated randomly with carefully designed noise and outlier distributions resembling real-world data. Generally speaking, a synthetic view-graph is generated by the following steps: 1) The number of nodes is sampled uniformly between $250$ and $1000$, and the orientation on each node is generated randomly on a horizontal plane. 2) Edges indicating relative rotations are randomly generated by the Erdős–Rényi model. The number of edges is set to $[10\%-30\%]$ of all possible pairs. 3) The relative orientations are corrupted by a Gaussian noise with a standard deviation $\sigma$ uniformly sampled in the range $[5^{\circ}-30^{\circ}]$. 4) Finally, $[0\%-30\%]$ of edges in the view-graph are replaced by random orientations as outliers. Similar to NeuRoRA \cite{ECCV2020neurora}, we generate $1,000$ view-graphs for training, $100$ for validation, and $100$ for testing. The parameters that yield the minimum validation loss are kept for testing.

\subtitle{Real-world datasets} We also evaluate on the real-world datasets \textit{1DSfM} \cite{wilson_eccv2014_1dsfm} and \textit{YFCC100} \cite{heinly2015_reconstructing_the_world}. 
The \textit{1DSfM} contains $14$ outdoor scenes with ground-truth camera poses and relative orientations computed by Bundler~\cite{schoenberger2016sfm}. Only the cameras with ground-truth orientations are used for training and testing.  The \textit{YFCC100} dataset consists of internet images at 72 city-scale scenes. We use the author's reconstructed camera poses as ground-truth and use relative orientations by COLMAP \cite{schoenberger2016sfm} provided in MSP\cite{Yang2021CVPR} for training and testing.
We train and test our method on the \textit{1DSfM} dataset in a leave-one-out manner. The \textit{YFCC100} dataset is split into two sets, one for training ($58$ scenes) while the other for testing ($14$ scenes).
To avoid overfitting, we randomly drop $20\%$ edges of the view-graph during training.
Due to the limitation of the training sample, we use CleanNet-SPT \cite{ECCV2020neurora} for global camera orientation initialization during training and testing on \textit{1DSfM} and \textit{YFCC100}.

\subtitle{Comparison} We compare our method with conventional optimized-based methods, including Chatterjee \etal \cite{ICCV2013}, MPLS \cite{Shi2020MessagePL}, Arrigoni \etal \cite{Arrigoni2018RobustSI}, Wang \etal \cite{wang2013exact}, Weiszfeld \cite{CVPR2011hartley}, Shonan~\cite{dellaert2020shonan}, MAKS~\cite{moreira2021rotation} and state-of-art deep learning based methods, including NeuRoRA \cite{ECCV2020neurora}, MSP \cite{Yang2021CVPR}. We use a publicly available evaluation script~\cite{ICCV2013} to compare predicted absolute camera orientations and ground-truth camera orientations in terms of mean(mn) and median(md) angular errors. Notice that RAGO does not resolve the gauge ambiguity. The output camera rotations need to align with the ground truth to evaluate the accuracy.

\subtitle{implementation details}
Our approach is implemented in Pytorch with an Nvidia V100 GPU. The model is trained with a adamW \cite{loshchilov2017decoupled} optimizer ($\beta_{1}=0.9,\beta_{1}=0.999$). The training runs for $20,000$ epochs started with a learning rate $1\times 10^{-3}$. After $100$ epochs, the learning rate decay exponentially by $0.999$ for each epoch. The $\gamma$ defined in \secref{sec:training_loss} is set as $0.8$ for all experiments.
During Training, we set $t_g$ to $3$, $t_e$ to $1$, and $t_n$ to $4$. We empirically terminate it when $t_g=5$ during testing.
The channel number of feature and hidden state is $48$.  
We use the Orth6D rotation representation proposed in~\cite{zhou2019continuity}. Orth6D is a continuous 6D space for 3D rotation matrices, while the quaternion, rotation matrix, and Euler angles are not contiguous in Euclidean space. In comparison, Orth6D enables RAGO to use the rotation matrix as the neural network's direct input and output. 
Please refer to supplementary material for more details.
\begin{table}[!t]
  \begin{center}
    \scalebox{0.8}{
\begin{tabular}{lccccccc}
\hline
 \multirow{2}{*}{Method} & \multicolumn{2}{c}{$t=1$} &\multicolumn{2}{c}{$t=5$} & \multicolumn{2}{c}{opt.} & Converge \\
  & mn & md & mn  & md & mn & md & Y/N \\
\hline\hline
\underline{Ours} & \textbf{4.03} & \textbf{2.41}  & \textbf{0.66} & \textbf{0.20} & \textbf{0.24} & \textbf{0.04} & Y\\
NeuRoRA\cite{ECCV2020neurora} & - & -  & - & - & 1.35 & 0.65 & Y\\
Chatterjee\cite{ICCV2013} & - & - & - & - & 2.20 & 1.30 & Y\\
Shonan\cite{dellaert2020shonan}& - & - & - & - & 2.43 & 1.58 & Y\\
MAKS\cite{moreira2021rotation}& - & - & - & - & 2.64 & 1.40 & Y\\
Wang\cite{wang2013exact} & - & - & - & - & 2.77 & 1.40 & Y\\
Arrigoni\cite{Arrigoni2018RobustSI} & - & - & - & - & 2.92 & 1.42 & Y\\
Weiszfeld\cite{CVPR2011hartley} & - & - & - & - & 3.35 & 1.02 & Y\\
\hline
\hline
\underline{with GRU} & 4.03 & 2.41  & \textbf{0.66} & \textbf{0.20} & \textbf{0.24} & \textbf{0.04} & Y\\
w/o GRU &  \textbf{3.31} & \textbf{2.18}  & 0.75 & 0.26 & 0.46 & 0.17 & N\\
\hline
\underline{SRA Cost} & \textbf{4.03} & \textbf{2.41}  & 0.66 & 0.20 & \textbf{0.24} & \textbf{0.04} & Y\\
Deg. Met. & 4.38 & 2.99  & \textbf{0.48} & \textbf{0.13} & 0.28 & 0.06 & Y\\
Null Vec. & 4.40 & 3.71  & 1.09 & 0.46 & 0.65 & 0.32 & Y\\
MRA Cost & 6.65 & 4.23  & 3.95  & 2.31 & 3.80 & 2.22 & N\\
\hline
\underline{3 Edge Conv} & \textbf{4.03} & \textbf{2.41}  & \textbf{0.66} & \textbf{0.20} & \textbf{0.24} & \textbf{0.04} & Y\\
2 Edge Conv & 4.37 & 2.62  & 0.68 & 0.21 & 0.41 & 0.11 & Y\\
1 Edge Conv & 7.15 & 4.05  & 1.39 & 0.58 & 0.71 & 0.25 & Y\\
\hline
\underline{Random Init.} & 4.03 & 2.41  & 0.66 & 0.20 & 0.24 & \textbf{0.04} & Y\\
Rand. SPT & 3.24 & 1.84  & 0.44 & 0.14 & 0.27 & 0.06 & Y\\
Clean. SPT & \textbf{2.74} & \textbf{1.44}  & \textbf{0.34} & \textbf{0.12} & \textbf{0.23} & \textbf{0.04} & Y\\
\end{tabular}
}
  \end{center}

  \caption{The results of comparison and ablation study on the synthetic dataset. We mark the final results of various MRA methods \cite{ICCV2013,wang2013exact,ECCV2020neurora,CVPR2011hartley,Arrigoni2018RobustSI,moreira2021rotation,dellaert2020shonan} as opt. The average mean(mn) and median(md) angular errors on the view-graphs in the test set are reported. The entries with the best performance are \textbf{bolded}. The settings used in the proposed model are \underline{underlined}.}
\label{table:ablation_table}
\vspace{-1em}
\end{table}
\begin{table*}[!t]
\begin{center}
\scalebox{0.85}{
\begin{tabular}{ccl|cc|cc|cc|cc|cc|cc}
\hline
\multicolumn{3}{c|}{Datasets} & \multicolumn{2}{c|}{Chatterjee\cite{ICCV2013}} & \multicolumn{2}{c|}{Weiszfeld\cite{CVPR2011hartley}}& \multicolumn{2}{c|}{NeuRoRA\cite{ECCV2020neurora}} & \multicolumn{2}{c|}{MPLS\cite{Shi2020MessagePL}}&\multicolumn{2}{c|}{MSP\cite{Yang2021CVPR}} & \multicolumn{2}{c}{Ours}  \\
\hline \hline
\#image & \#edge & Names & mn    & md    & mn    & md  & mn    & md  & mn   & md   & mn   & md& mn   & md   \\ 
                  577  & 59.5\%  & Alamo               & 4.2  & 1.1  & 4.9 &1.4 & 4.94          & 1.16          & 3.44                 & 1.16     & \underline{2.89}                      & \underline{1.07}    & \textbf{2.82}      &   \textbf{0.88}                         \\
    227  & 66.8\%  & Ellis Island       & 2.8  & 0.5& 4.4  & 1.0 & 2.59          & 0.64 & 2.61                 & 0.88    & \underline{1.88}                   & \underline{0.83}          & \textbf{1.74}&  \textbf{0.46}                   \\
677  & 17.5\% & Gendarmenmarkt  & 37.6 & 7.7 & 29.4 & 9.6 & \textbf{4.51}          & \underline{2.94}          & 44.9                 & 8.0     & 6.29  & 3.69     & \underline{5.24} &     \textbf{2.68}        \\
341  & 30.7\%  & Madrid Metropolis              & 6.9  & 1.2 & 7.5  & 2.7 & \textbf{2.55} & 1.13          & 4.65                 & 1.26      & \underline{2.96}                              & \underline{1.09}  & 3.05& \textbf{1.03}                    \\
450  & 46.8\%  & Montreal Notre Dame & 1.5  & \underline{0.5}& 2.1 & 0.7 & 1.2           & 0.6           & 1.04        & 0.51      & \underline{0.91}           & \underline{0.5}     &\textbf{0.86}&\textbf{0.46}            \\
338  & 39.5\%  & Piazza del Popolo   & 4    & 0.8& 4.8   & 1.3 & 3.05          & 0.79          & 3.73                 & 1.93          & \underline{2.68}                   & \underline{0.76}      &\textbf{1.91}&\textbf{0.63}         \\
  1084 & 10.9\%  & Roman Forum         & 3.1  & 1.5 & 4.8  &1.8 & \underline{2.39}          & 1.31          & 2.62                 & 1.37    & \textbf{2.04}                     & \underline{1.19}   &2.55& \textbf{1.10}                 \\
472  & 18.5\%  & Tower of London        & 3.9  & 2.4& 4.7 & 2.9 & 2.63          & 1.46          & 3.16                 & 2.2           & \underline{2.55}                     & \underline{1.25}      &\textbf{2.51}&\textbf{1.20}         \\
5058 & 4.6\%   & Trafalgar           & \underline{3.5}  & \underline{2} & 15.6  & 11.3   &5.33        & 2.25          &  \multicolumn{2}{c|}{-} & \multicolumn{2}{c|}{\footnotesize{Out of Memory}} & \textbf{2.23}& \textbf{1.53} \\ 
789  & 5.9\%   & Union Square       & 9.3  & 3.9 & 40.9 & 10.3 & 5.98          & 2.01          & 6.54                 & 3.48     & \textbf{4.37}                     & \textbf{1.85}           &\underline{4.68}& \underline{1.92}        \\
836  & 24.6\%  & Vienna Cathedral    & 8.2  & 1.2& 11.7  & 1.9 & \textbf{3.9}  & 1.5           & 7.21                 & 2.83    &       \underline{3.91}                              & \underline{1.1}       &6.05&    	\textbf{0.89}    \\
437  & 26.5\%  & Yorkminster         & 3.5  & 1.6& 5.7  & 2.0 & 2.52          & 0.99          & 2.47                 & 1.45     & \underline{2.27}                    & \textbf{0.91}         & \textbf{2.18}& \underline{0.92}          \\
2152 & 10.2\%  & Piccadilly          & 6.9  & 2.9& 26.4  & 7.5 & 4.75          & 1.91          & 3.93                 & 1.81        & \underline{3.63}                    & \underline{1.8}   &\textbf{2.44}&\textbf{0.58}               \\
332  & 29.3\% & NYC Library        & 3    & 1.3   & 3.8    & 2.1 & \underline{1.9}           & 1.18          & 2.63                 & 1.24     & \textbf{1.75}                    & \underline{1.12}   &2.02& \textbf{0.71}          \\
\hline    
\end{tabular}
}
\end{center}
\vspace{-0.5em}
\caption{Comparison of results on the \textit{\textbf{1DSfM}} dataset. We compare our method with various SOTA MRA methods. Mean(mn) and median(md) angular errors on the estimated absolute rotations are compared. The entries with the best performance is \textbf{bolded}, the second is \underline{underlined}. Notice that the result of MSP is computed by using additional input, such as image and correspondence.}
\label{table:1dsfm_table}
\end{table*}

\subsection{Synthetic dataset}

\tabref{table:ablation_table} shows the results of our method on the synthetic dataset. We show the average mean and median angular errors for all view-graphs in the test set. We report the results of our method at iteration $1$ and $5$. The result of iteration $20$ is marked as opt. We compare our method with the learning-based approach NeuRoRA \cite{ECCV2020neurora} and the conventional method Chatterjee \cite{ICCV2013}, Arrigoni \cite{Arrigoni2018RobustSI}, Wang \cite{wang2013exact} and Weiszfeld \cite{CVPR2011hartley}, Shonan~\cite{dellaert2020shonan} and MAKS~\cite{moreira2021rotation}. The final results of these approaches are marked bolded. Although our results have relatively large error at the 1st iteration, they outperform the other methods after the $5$-th iteration and finally converge to $0.24$ and $0.04$ in mean and median angular error. \figref{fig:syn_mean} shows the mean error during iterative optimization, indicating fast convergence and significantly improved accuracy compared with conventional and learning-based methods.

\subsection{Ablation study}
To understand the different components of our method, we conduct an ablation study on the synthetic dataset, with results summarized in \tabref{table:ablation_table}.

\subtitle{GRU Module} To see the effectiveness of utilizing the history information during optimization, we replace the GRU module with a $6$-layers MLP. 
The errors also reduce rapidly in the first several iterations, but the results eventually diverge, yielding a poorer accuracy.

\subtitle{Different Metrics on Cost Graph} To evaluate the effectiveness of different cost functions in the cost graph, we experiment the cost defined in \secref{sec:define_cost} with angular degree distance, MRA cost, and null vector. 
For angular degree distance cost, we substitute angular degree error for L1 norm.
For MRA cost, we maintain the cost constant on edge but substitute the cost on node with the cost function of MRA defined as \equref{eq:mra_objective}.
Finally, to assess the efficacy of the cost graph, we substitute a zero vector as a null vector for all costs on the graph.
As shown on \tabref{table:ablation_table}, the model with the angular degree metrics has results as the proposed model. Although the cost graph with the null vector has inferior results than the proposed model, it still outperforms the baselines due to iterative optimization and temporal information. The result drops dramatically, if we use MRA cost because it is evaluated on the whole view-graph, making it hard for the neural network to learn. This comparison demonstrates the advantages of solving MRA through solving multiple SRA problems.

\subtitle{Number of Edge Conv} As introduced in \secref{sec:feat_ext}, At each Edge Convolution layer, the node and edge features are updated by aggregating the neighbors' information and then passed to the next layer. Thus, the number of layers of Edge Convolution layer will affect the receptive field of each entity on the view graph. To demonstrate the effectiveness of the size of the receptive field, We train $3$ models with different numbers of Edge Convolution layers in MPNN $\Theta_{\text{cost}}$. The models trained with $2$ and $3$ Edge Convolution layers have comparable results, while the performance will drop significantly on the setting with only $1$ Edge Convolution layer. This experiment shows the information from farther neighbors would be helpful to achieve a better convergence. 

\subtitle{Camera Orientation Initialization} To study the effectiveness of different initialization, we train $3$ models respectively with: random initialization (Random Init.), random spanning tree initialization (Rand. SPT), CleanNet-SPT initialization (Clean. SPT). For the random spanning tree initialization, we randomly generate a spanning tree of the view-graph, then uniformly select a root node to compute propagate an initialization through the spanning tree. CleanNet-SPT uses an MPNN with 3 Edge Convolution layers to predict each edge is an outlier or not. Then the node with the most neighbors would be chosen as the root node to propagate an initialization through the minimum spanning tree.  The ablation result shows that all three initialization methods can converge to an optimized solution, while the models with random SPT and CleanNet-SPT initialization could converge faster. Notice that RAGO has not been associated with gauge ambiguity. The strategy of how to select a root node becomes trivial. Although our method has similar optimized results with the different approaches on the synthetic dataset, an appropriate initialization is still needed for more complicated real-world datasets due to the limitation of the training samples.


\begin{table*}[h!t]
\begin{center}
\scalebox{0.9}{
\begin{tabular}{ccl|cc|cc|cc|cc|cc}
\hline
\multicolumn{3}{c|}{Datasets}    & \multicolumn{2}{c|}{Chatterjee\cite{ICCV2013}} & \multicolumn{2}{c|}{NeuRoRA\cite{ECCV2020neurora}} & \multicolumn{2}{c|}{\begin{tabular}[c]{@{}c@{}}NeuRoRA~\cite{ECCV2020neurora}\\+ MSP Refine.\cite{Yang2021CVPR}\end{tabular}} & \multicolumn{2}{c|}{MSP\cite{Yang2021CVPR}} & \multicolumn{2}{c}{Ours} \\
\hline \hline
\#image &\#edges  & Names                  & mn                        & md          & mn                        & md                      & mn                          & md                 & mn                       & md                       & mn                       & md                     \\
1881      & 4.2\%  & colosseum\_exterior                  & 7.21                      & 4.16                    & 25.09                       & 2.48        & 22.81                      & 3.06         & \underline{2.7}           & \underline{1.66}              &\textbf{1.97} & \textbf{1.35}    \\
228    & 28.2\%    & piazza\_san\_marco               & 2.31                      &\underline{1.2}                  & 8.08                        & 4.01           & 3.53                         & 2.28     & \textbf{2.02}            & 1.55                     &\underline{2.24}& \textbf{1.12}   \\
163      & 66.6\%  & big\_ben\_2                         & 14.56                     & 2.96                    & 7.56                        & 2.36        & \underline{5.58}                       & \underline{1.38}       & 5.77            & 1.43         &\textbf{3.42}&\textbf{1.21} \\
182     & 41.7\%   & palazzo\_pubblico                   & 5.69                      & 1.91                    & 3.58                        & 1.58       & 3.49                        & \underline{1.21}         & \underline{3.22}          & 1.4            &\textbf{2.16}& \textbf{0.91}\\
624      & 12.4\%  & louvre                             & 7.69                      & 2.69                    & 8.55                        & 4.82 & 6.48                       & 1.47               & \underline{5.04}             & \textbf{0.9}           &\textbf{3.47}&\textbf{0.90}  \\
188     & 59.5\%   & big\_ben\_1          & 12.57                     & 2.59                    & 5.22                        & 2.60            & 9.01                      & 2.60     & \underline{3.42}           & \underline{1}     &\textbf{2.97}&\textbf{0.94}    \\
104    & 63.2\%    & petra\_jordan                & 8.68                      & 1.76                    & 5.15                        & 3.19       & 4.19                      & 0.75      & \underline{2.85}            & \textbf{0.5}         &\textbf{2.76}& \underline{0.81}    \\
100   & 53.7\%    & statue\_of\_liberty\_2         & 10.06                     & 4.17                    & 5.80                        & 2.23        & 4.90                      & 1.99         & \underline{2.93}            & \underline{1.2}      &\textbf{2.54}&\textbf{1.02}         \\
269    & 23.1\%   & st\_peters\_basilica\_interior\_2     & 7.43                      & 2.72                    & 6.24                       & 2.44         & 4.91                      & \underline{ 1.08 }       &\underline{4.63}            & 1.43       &\textbf{3.33}&\textbf{0.92}    \\
90     & 66.0\%    & statue\_of\_liberty\_1           & 6.79                      & 2.45                    & 5.71                        & 2.34    & 4.43                       & 1.99             & \textbf{3.22}            & \textbf{1.35}       &\underline{3.44}&\underline{1.55}       \\
103    & 55.9\%    & florence\_cathedral\_side       & 8.56                      & 3.46                    & 2.87                        & 1.19     & 2.91                       & \underline{0.78}           & \textbf{1.55}            & \textbf{0.62}         &\underline{1.75}&1.57             \\
136    & 43.6\%    & palace\_of\_versailles\_chapel      & 13                        & 2.76                    & \underline{2.98}               & 0.96       &5.01              &1.13         & 3.12                     & \underline{0.64}    &\textbf{2.74}&\textbf{0.61}           \\
496    & 14.6\%    & notre\_dame\_rosary\_window   & 7.41                      & 1.94                    & 7.06                        & 3.83     & 4.41                       & 1.67           &\underline{2.79}            & \underline{0.96}     &\textbf{2.08}&\textbf{0.80}          \\
745   & 10.8\%    &  lincoln\_memorial\_statue       & 8.08                      & 1.54                    & 2.87                        & 1.48            & 3.74                        &1.19    & \underline{1.95}            & \textbf{0.96}    &\textbf{1.87}&\underline{1.21}             \\
 \hline    
\end{tabular}
}
\end{center}
\caption{Comparison of results on the \textit{\textbf{YFCC100}} dataset. We compare our method with various SOTA MRA methods, mean(mn) and median(md) angular errors on the estimated absolute rotations are compared. The entries with the best performance are bolded. The second is \underline{underlined}. Notice that the result of MSP is computed by using additional input, such as image and correspondence.}
\label{table:yfcc_table}
\vspace{-1em}
\end{table*}

\subsection{Results on Real World Dataset}
\subtitle{\textit{1DSfM}} The comparison on the \textit{1DSfM} dataset are listed in \tabref{table:1dsfm_table}. Notice that MSP \cite{Yang2021CVPR} uses additional information as input, \eg correspondences, while others only use observed relative orientations as input. Our method outperforms other methods in median angular error in most scenes. In terms of mean of angular error, our approach has the best performance on half of the scenes and yields comparable results on the remaining ones. Our graph optimizer only performs slightly inferior on \texttt{Gendarmenmarkt}, \texttt{Vienna Cathedral} and \texttt{Madrid Metropolis} compared with NeuRoRA \cite{ECCV2020neurora} with the same input.

\subtitle{\textit{YFCC100}} The results on \textit{YFCC100} dataset are listed on \tabref{table:yfcc_table}. We cite the result from MSP~\cite{Yang2021CVPR} directly. We compare our graph optimizer with several MRA methods. Our method outperforms previous methods in most scenes in terms of mean of angular error except \texttt{Florence\_cathedral\_side} because \textit{YFCC100} contains more view-graphs for training compared with \textit{1DSfM}. NeuRoRA produces large mean and median error on \texttt{colosseum\_exterior}, while ours could still optimize to a strong result. MSP\cite{Yang2021CVPR} has a slightly better result on \texttt{statue\_of\_liberty\_1} compared with ours due to additional input and robust global camera initialization.

\begin{table}[!t]
  \begin{center}
     \scalebox{0.77}{
\begin{tabular}{c|c||c|cc||c|cc||c}
  \hline
  Robust. & Eval. & Train. & mn & md & Train. & mn & md & Time \\
  \hline
  \hline
  \multirow{3}{*}{$|\mathcal{V}|$} & 300  & \multirow{3}{*}{$600$} & 0.27 & 0.04 & 300 & 0.24 & 0.04 & 0.005 \\
   & 600 &  & 0.15 & 0.03 & 600 & - & - & 0.007 \\
   & 1500 &  & 0.38 & 0.16 & 1500 & 0.24 & 0.09 & 0.011 \\
  \hline
  \multirow{3}{*}{$|\mathcal{E}|$} & $3\%$ & \multirow{3}{*}{$30\%$} & 2.98 & 0.47 & $3\%$ & 1.75 & 0.32 & 0.006 \\
   & $30\%$ &  & 0.15 & 0.03 & $30\%$ & - & - & 0.007 \\
   & $60\%$ &  & 0.35 & 0.18 & $60\%$& 0.14 & 0.04 & 0.015 \\
  \hline
  \multirow{3}{*}{$\sigma$} & $5^{\circ}$ & \multirow{3}{*}{$15^{\circ}$} & 0.13& 0.03 & $5^{\circ}$ &  0.17& 0.05& 0.008 \\
   & $15^{\circ}$ &  & 0.15 & 0.03 & $15^{\circ}$ & - & - & 0.007 \\
   & $55^{\circ}$ &  & 0.57 & 0.33 & $55^{\circ}$ & 0.35 & 0.17 & 0.008  \\
  \hline
  \multirow{3}{*}{$o$} & $3\%$ & \multirow{3}{*}{$15\%$} & 0.06 & 0.04 & $3\%$ & 0.07 & 0.04 & 0.011 \\
   & $15\%$ &  & 0.15 & 0.03 & $15\%$ & - & - & 0.007 \\
   & $60\%$ &  & 0.82& 0.13 & $60\%$ & 0.53 & 0.12 & 0.011 \\
\end{tabular}
}
  \end{center}
  \caption{The results of robustness check on the different synthetic datasets. We compare the results of our models trained on the synthetic datasets with the different settings. We report the result in terms of average mean and median angular error. We show the runtime of our graph optimizer with different kinds of view-graph in terms of second per iteration.}
\label{table:robust_table}
\end{table}
\subsection{Robust Check}
\label{sec:robust}

This experiment shows the generalization capacity of the RAGO. To generate the synthetic datasets, we use the configuration of $(|\mathcal{V}|, |\mathcal{E}|, \sigma, o)=(600,30\%,15^{\circ}, 15\%)$ as the default setting. To check the individual effects of different sensor settings, we generate some synthetic datasets varying: 1) the number of the cameras $|\mathcal{V}|$, 2) the percentage of the edges $|\mathcal{E}|$, 3) std of the edge error $\sigma$ and 4) the percentage of outlier edge $o$. RAGO is then trained on one of such datasets and evaluated on the others. Each dataset consists of $1,000$ view-graphs for training, $100$ for testing.The results of the robustness check as shown in \tabref{table:robust_table}. We report the average mean and median angular error on the testing set.
For the column from $3$ to $5$, We train RAGO on the default synthetic dataset and evaluate it under different configurations. The RAGO training and testing results under the synthetic dataset of the same settings are shown from column $6$ to $8$. 
The experiments demonstrate that RAGO generalizes well across dataset changes except when the model is trained on the sparse view-graph.

\subsection{Time-Space Complexity and Model Size}
\label{sec:timeandmem}

RAGO only contains appropriately $0.396$M parameters and can be easily deployed for real-world applications. We deploy our method on a Nvidia V100 GPU and evaluate it using view-graphs with different of nodes and edges. The average running time for each iteration during the optimization is shown on \tabref{table:robust_table}. For the view-graph with $600$ nodes and $30\%$ edges, RAGO only takes $0.007$ seconds for each iteration. It uses $0.015$ seconds per iteration on the view-graph with $600$ nodes and $60\%$ edges. The time and space complexity of our graph optimizer is $O(|\mathcal{E}|+|\mathcal{V}|)$.
The running time and memory consumption of our method increases linearly as the size of the input view-graph increases.


\section{Conclusion}
\label{sec:conclusion}
We  propose a learning-to-optimize graph-based optimizer (RAGO) for the Multiple Rotation Averaging (MRA) problem.
RAGO solves the original MRA by building a cost graph based on the Single Rotation Averaging (SRA) objective function to update camera orientations iteratively.
During optimization, the relative orientations are rectified to handle the outliers and the noises.
The Gated Recurrent Unit (GRU) is employed to exploit temporal information during iterations.
RAGO outperforms previous methods on synthetic and real-world datasets and is also highly efficient in running time and memory.

\label{sec:limitation}
\subtitle{Limitation} RAGO belongs to learning-based methods, which suffer from cross-domain generalization problems, e.g., RAGO trained on indoor scenes might work poorly on outdoor scenes. Furthermore, RAGO is trained in a supervised manner, while precise ground truth of real data is hard to obtain. We leave unsupervised training for future work.

{\noindent}{\textbf{Acknowledgement.}} This research is supported in part by the Canada NSERC Discovery project 611664 and the National Natural Science Foundation of China (NSFC), under grants No.~61872067, No.~61720106004 and  No.~62102356.





\bibliographystyle{ieee_fullname}
\bibliography{RAGO_camera_ready}

\end{document}